\documentclass[runningheads]{llncs}
\usepackage{graphicx}
\usepackage{physics,amsmath}
\usepackage{multirow}
\usepackage{svg}
\begin{document}
\title{EEML: Ensemble Embedded Meta-learning\thanks{Supported by Harbin Institute of Technology.}}
%
%
\author{Geng Li\inst{1}\orcidID{0000-0002-9733-6891} \and
Boyuan Ren\inst{2}\orcidID{0000-0002-1291-5377} \and
Hongzhi Wang\inst{3}\orcidID{0000-0002-7521-2871}}
%
\institute{
Harbin Institute of Technology, Harbin, CHINA 
\email{21S003022@stu.hit.edu.cn } \and
Harbin Institute of Technology, Harbin, CHINA 
\email{1180300207@stu.hit.edu.cn } \and
Harbin Institute of Technology, Harbin, CHINA 
\email{wangzh@hit.edu.cn }
}
\maketitle              
\begin{abstract}
To accelerate learning process with few samples, meta-learning resorts to prior knowledge from previous tasks. However, the inconsistent task distribution and heterogeneity is hard to be handled through a global sharing model initialization. In this paper, based on gradient-based meta-learning, we propose an ensemble embedded meta-learning algorithm (EEML) that explicitly utilizes multi-model-ensemble to organize prior knowledge into diverse specific experts. We rely on a task embedding cluster mechanism to deliver diverse tasks to matching experts in training process and instruct how experts collaborate in test phase. As a result, the multi experts can focus on their own area of expertise and cooperate in upcoming task to solve the task heterogeneity. The experimental results show that the proposed method outperforms recent state-of-the-arts easily in few-shot learning problem, which validates the importance of differentiation and cooperation. 

\end{abstract}
\section{Introduction}

One of the Intelligent advantages of human being is acquiring new skill for unseen task from few samples. To achieve the fast learning capability, meta-learning also known as learning to learn, leverages what we learned from past tasks to improve learning in a new task. As a common practice for few-shot learning problem, meta-learning contains several series which differ in the way of employing prior knowledge \cite{survey}: metric-based \cite{metric1,metric2}, model-based \cite{network1}, gradient-based(a.k.a. optimization-based) meta-learning \cite{maml}. \\

Though these methods earn relative improvement comparing with past methods, most of them hypothesize the prior knowledge learned from past should be shared indifferently for all tasks. As a consequence, they suffer from the tasks heterogeneity. On the other hand, a few research works try to fix the problem by customizing the prior knowledge to each task \cite{l2f,leo,hsml}. However, the defect of such methods lie in resorting to only a single expert who may not be sufficient for achieving the best result. In other words, instead of reasonably utilizing overall intelligence, these methods prefer to choose an expert as  dictator.\\

Hence, we are motivated to pursue a meta-learning framework both effectively diverge knowledge into multi experts and comprehensively utilizing overall experts to aggregate an stable answer for new task to enhance the final performance. The inspiration comes from classical ensemble learning, bagging \cite{bagging} or random forest \cite{random_forest}. In ensemble learning, though single learner may perform trivially or sometimes blindly on the whole task distribution, but the ensemble mechanism can help each expert reach a relatively good cooperative result through counteracting mutual bias. \\

Inspired by this, we propose a novel meta-learning algorithm named Ensemble Embedded Meta-learning(EEML). The key idea of EEML is to organize heterogeneous training tasks into matching experts and collecting all experts opinions to form a comprehensive answer for a new task. In this paper, we resorts to gradient-based meta-learning as the backbone which instantiates knowledge as parameter initialization. Specifically, we employ a task embedding cluster to cover the domain of each expert. The gradient of task would be considered as a strong embedding and we assign training task to the matching expert according to the similarity between the gradient and the expert domain. In predicting phase, all experts can utilize few samples to contribute its opinions, and the final result is composed as a weighted voting answer from all experts. \\

We would highlight the contributions of EEML as two points: 1) it employs multi-experts cooperation framework instead of single expert for uncertain tasks, so that it outperforms recent state-of-the-art meta-learning algorithms in few-shot learning. 2) as we know, EEML is the first generic meta-learning  algorithm combined with ensemble learning, whose experiment results show great confidence in exploring further in composing meta-learning and ensemble learning for future work.

\section{Related Work}
\subsection{Meta-learning}
Meta-learning is understood as “learning to learn” \cite{old3,old4,old5,old6}, it starts with the form of learning to optimize model \cite{old1,old2} and now are mainly divided into three series, metric-based meta-learning \cite{metric1,metric2,metric3,metric4,metric5}, network-based meta-learning \cite{network1,network2,network3,network4,network5} and gradient-based meta-learning \cite{gbml1,gbml2,gbml3,gbml4,gbml5,gbml6}. The goal of metric-based algorithms is to measure the similarity between support samples and query samples with elaborate metric methodology. Network-based approaches emphasize finding structures effective in capturing the prior knowledge in training data. Gradient-based meta-learning aims to use different task gradients to optimize a base model with a batch form.

In this work, we focus on the discuss about gradient-based meta-learning. The most powerful representation should be MAML \cite{maml}, which presents the “inner-loop and outer-loop” framework commonly used by plenty of followers. Reptile \cite{reptile} uses accumulated first gradient replace the second derivative and earns speeding up but instability. MetaOptNet \cite{metaoptnet} and \cite{metric2,cifarfs_dataset} reset the inner optimization by dividing the model into feature extractor and head classifier and apply on complex base learner like ResNet-12 without overfitting. LEO \cite{leo} introduces lower-dimensional latent space to consider task-conditioned initialization. L2F \cite{l2f} uses extra MLP network to help attenuate the base learner parameters to gain a task-conditioned initialization. 
\subsection{Ensemble Learning}
Ensemble learning exploits multiple base machine learning algorithms to obtain weak predictive results and fuse results with voting mechanism to achieve better performance than that obtained from any constituent algorithm alone \cite{ensemble_learning}. The main classical ensemble learning algorithms are AdaBoost \cite{adaboost}, Random Forest \cite{random_forest} and Bagging \cite{bagging}. In this paper, to start with clarity we mainly rely on the bagging mechanism which is simple and efficient, besides, it's naturally adaptive to parallel training paradigm, which can effectively speedup our training process. However, with the precisely designed coefficients adopted during training and test phase, our method has inherent difference with original bagging.

\section{Preliminaries}
\subsection{The Problem Definition}
Supposed that a sequence of tasks $\{\mathcal{T}_1,\mathcal{T}_2,...,\mathcal{T}_N\} $ are sampled from a probability $p(\mathcal{T})$. Each task sampled from $p(\mathcal{T})$ is defined as $\{x_{i},y_{i}\}^{m}_{i=1}$. Dividing parts samples into training set $ \mathcal{D}^{tr}=\{x_i,y_i\}^{n^{tr}}_{i=1}$ and the rest as test set $ \mathcal{D}^{te} =\{x_i,y_i\}^{m}_{i=n^{tr}+1}$, we then can also describe a task $\mathcal{T}$ as $\{\mathcal{D}^{tr},\mathcal{D}^{te}\}$. Given a model $f_{\theta}$, the training aim is to obtain the optimal parameters $\theta$ to minimize the loss function:
$$
 {Loss(f_{\theta}(D^{tr},\{x_i\}^{m}_{i=n^{tr}+1}),\{y_i\}^{m}_{i=n^{tr}+1}))}   
$$
which can be simply noted as $\mathcal{L}_{\mathcal{T}}^{\mathcal{D}^{te}}(f_{\theta})$. 

Based on this, the meta-learning considers model $f$ as a learning process, which earn $\theta^{\prime}$ from $\theta$ and give final prediction based on $\theta^{\prime}$. To find a globally shared parameters $\theta_g$, we follow: 
\begin{equation}
\min_{\theta_g} \Sigma^{N}_{k=1} Loss(f_{\theta_g}(\mathcal{D}^{tr}_k,\{x_{k,i}\}^{m}_{i=n^{tr}+1}),\{y_{k,i}\}^{m}_{i=n^{tr}+1}))
\end{equation}
Different meta-learning algorithms mainly diverge in $f$, which is also utilized in test phase to map $\theta_g$ to $\theta^{\prime}_g$ for new task $t_i$  and gain predictions corresponding.

\subsection{The Gradient-based Meta-learning}
Here, we compactly give an overview of the representative algorithm of gradient-based meta-learning,i.e. model-agnostic meta-learning. With the aim of searching for a globally sharing parameters initialization for unseen tasks, MAML define learning process $f$ like below. Formally, given the task distribution $p(\mathcal{T})$, we sample a task $\mathcal{T}_{i}$ from it and define the base learner as $g_\theta$ then we can get:
\begin{equation}
\theta_{i}^{\prime}=\theta- \nabla_{\theta} \mathcal{L}_{\mathcal{T}_{i}}^{\mathcal{D}_{i}^{tr}}\left(g_{\theta}\right)    
\label{inner_loop}
\end{equation}

\begin{equation}
f_\theta(\mathcal{D}_{i}^{tr},\{x_i\}^{m}_{i=n^{tr}+1}) = g_{\theta_{i}^{\prime}}(\{x_i\}^{m}_{i=n^{tr}+1})
\label{ffunction}
\end{equation}
To update $\theta$, MAML then follows:
\begin{equation}
\theta \leftarrow \theta- \nabla_{\theta}\sum_{\mathcal{T}_{i} \sim p(\mathcal{T})}\mathcal{L}_{\mathcal{T}_{i}}^{\mathcal{D}^{te}_{i}}\left(f_{\theta_{i}}\right)
\label{outer_loop}
\end{equation}
For simplicity of description, We define gradient of task $\mathcal{T}_i$,   $\nabla_{\theta} \mathcal{L}_{\mathcal{T}_{i}}^{\mathcal{D}_{i}^{te}}\left(f_{\theta_{i}}\right)$, as $u_i$ in following  sections.

\begin{figure}[t]
\includegraphics[width=\textwidth]{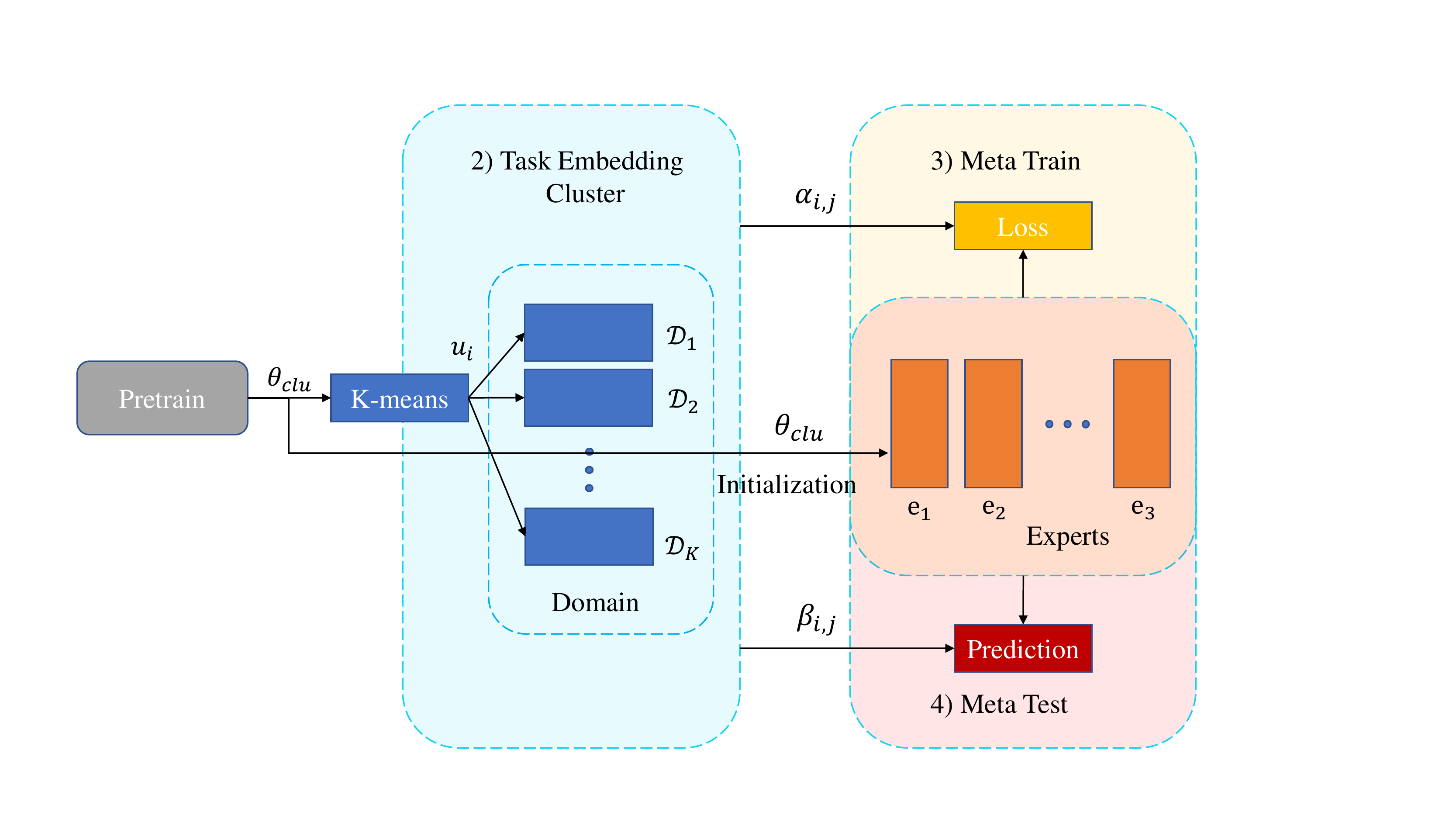}
\caption{The framework of EEML mainly concerns three stages: a)Task Embedding Cluster: We employ the $\theta_clu$ to gain task gradient embedding $u_i$, then leverage K-means to get Domains. b) Meta Train: via calculating coefficients $\alpha_{i,j}$, we train experts towards costumed loss function. c) Meta Test: Given coefficients $\beta_{i,j}$, we gain cooperative results through weighted voting mechanism. } \label{fig}
\end{figure}

\section{Methodology}
In this section, we detail the proposed EEML algorithm like the Fig.~\ref{fig}. The EEML aims to raise a group of experts covering heterogeneous knowledge domains in meta-training, then collect all experts opinions in a weighted voting schema to achieve better results in test phase. The following parts of this section are organized in task embedding cluster, meta-training and meta-test.

\subsection{Task Embedding Cluster}
Instead of abstracting task embedding from only labeled data, we argue that gradient $u_i$ calculated in (\ref{outer_loop}) can consider dynamical training process which implies optimization direction information ignored by static task embedding mechanism like auto-encoder or rnn-based encoder. With gradient $u_i$ considered as task $\mathcal{T}_i$ embedding, we resorts to a classical cluster algorithm, K-means to divide sampled tasks into different groups. The distance function leveraged in K-means is cosine distance which focuses on the difference in direction instead of embedding norm. Given hyperparameter $K$, we then obtain $K$ clusters which correspond to different optimization direction for specific model parameters $\theta_{clu}$. We collect the centers of different clusters as $\{c_1,c_2,...,c_K\}$. To measure the distance between future coming task and existing clusters, we follow the cosine distance as the standard metric. 

\subsection{Meta Train}
Unlike training single dictator model, raising a group of experts should carefully maintain the balance between diversity and generalization. To start with a generic initialization, we initialize the experts $\{g_{e_1},g_{e_2},..,g_{e_K}\}$ as $K$ copies from $g_{\theta_{clu}}$, whose generalization then can be controlled through parameter initialization $\theta_{clu}$. To get an effective initialization without introducing too many operations, we simply exploit MAML to earn a relatively proper $\theta_{clu}$ as the initialization parameters for clustering and experts. We can adjust the pretraining via limiting the total training epochs.\\

In training stage, to strengthen the diversity of experts in training paradigm, we utilize the loss function below as the target training loss. 

\begin{equation}
    \mathcal{L}oss = \Sigma_{i=1}^{N} \Sigma_{j=1}^{K} {\alpha_{i,j}} \mathcal{L}_{\mathcal{T}_i}^{\mathcal{D}^{te}_i}(g_{e_j})
    \label{total loss}
\end{equation}

The coefficient $\alpha_{i,j}$ would induce the final tendency of expert $e_i$ towards specific tasks. In our assumption, a qualified expert $e_i$ should be familiar with a task from a specific domain $D_i$ while has no strict demanding about the competence for other tasks outside. We define the $D_i$ as $\{\mathcal{T}_j | Dist(\mathcal{T}_j,c_i) = \min_{p=1:K} Dist(\mathcal{T}_j,c_p) \}$, which indicates a task $\mathcal{T}_j$ belonging to $D_i$  should be nearest to the domain center $c_i$ comparing with other domain centers. To expand the generalization, we do not assign coefficient $\alpha_{i,j}$ from the indicator of whether task $\mathcal{T}_i$ is in $D_j$, instead, we choose the cosine similarity metric after softmax layer as final $\alpha_{i,j}$ to ensure the probability sum as one. \\

In this paper, instead of considering static task embedding  without any information from model, we decide to leverage the gradient $u_i$ as the embedding of task $\mathcal{T}_i$. With both model optimization information and labelled data contained, the $u_i$ help EEML achieve better performance. 

\begin{equation}
    \begin{split}
    \vectorbold{\alpha}  &= Softmax(Sim(\mathcal{T}_j, \vectorbold{c})) \\
    &= Softmax  ( \frac{u_i \cdot \vectorbold{c}}{\Vert u_i \Vert \Vert \vectorbold{c} \Vert} )  
    \end{split}
\end{equation}

\subsection{Meta Test}
Facing unseen task from heterogeneity distribution, single dictator model is insufficient and easy to be blind. In our paradigm, all experts are able to join the test phase through two steps. First, the experts fine-tune the parameters with training set $\mathcal{D}^{tr}_i$ from $\mathcal{T}_{test}$. 

\begin{equation}
    e^{\prime}_i = e_i - \nabla_{\theta} \mathcal{L}_{\mathcal{T}_{test}}^{\mathcal{D}_{test}^{tr}}\left(e_i \right)    
    \label{alpha}
\end{equation}

With fine-tuned experts $\vectorbold{e}^{\prime} = {e^{\prime}_1,e^{\prime}_2,...,e^{\prime}_n}$. We leverage a weighted voting mechanism inspired by Bagging like:

\begin{equation}
    p_j = \Sigma_{i=1}^{K} \beta_{i,j} g_{e_i^{\prime}}(\mathcal{T}_j)
    \label{beta}
\end{equation}

The coefficient $\beta_{i,j}$ is designed as: 
\begin{equation}
    \beta_{i,j} = \frac{Sim(\mathcal{T}_j,c_i)}{err_{i,j}}  
\end{equation}

The $err_{i,j}$ is the error value of training set $\mathcal{D}_{j}^{tr}$ with model $g_{e_i}$, which implies the adaptation level of expert $e_i$. Like $\alpha_{i,j}$ we also take the softmax function as filter to maintain probability property. 

\begin{equation}
    err_{j} = Softmax(\mathcal{L}_{\mathcal{T}_j}^{\mathcal{D}^{tr}}(g_{\vectorbold{e}}))
\end{equation}

$\mathcal{L}_{\mathcal{T}_j}^{\mathcal{D}^{tr}}(g_{\vectorbold{e}})$ is a vector composed of $\{\mathcal{L}_{\mathcal{T}_j}^{\mathcal{D}^{tr}}(g_{e_1}),\mathcal{L}_{\mathcal{T}_j}^{\mathcal{D}^{tr}}(g_{e_2}),...,\mathcal{L}_{\mathcal{T}_j}^{\mathcal{D}^{tr}}(g_{e_K})\}$.
In our paradigm, the final prediction of all experts would be considered based on the  expert similarity with the task and the final performance. And
the dictator mechanism which consists of only one final model can be considered as a specific instance when $K=1$.

\section{Experiments}
In this section, we evaluate the effectiveness of our proposed EEML on both toy regression and complex image few-shot classification cases. This section is organized as Datasets and Implementation Details, Toy Regression, and Image Few-shot Classification.

\subsection{Datasets and Implementation Details}

\begin{table*}
\caption{Toy regression function family definition details.}
\label{tab:regression_definition}
\begin{tabular}{c|c|c|c|c|c}
\hline
\multirow{2}{*}{function family}  & \multirow{2}{*}{formulation ($y(x)$)}  &  \multicolumn{4}{c}{parameters range}\\
\cline{3-6}
& & $p_1$ & $p_2$ & $p_3$ & $p_4$\\
\hline
Sinusoids & $ p_1 \sin (p_2 x + p_3)$ & $[0.1,5.0]$ & $[0.8,1.2]$ & $[0,2 \pi]$ &  $-$ \\
Line & $p_1 x + p_2$ & $[-3.0,3.0]$ & $[-3.0,3.0]$ & $-$ & $-$\\
Quadratic & $p_1 x^2 + p_2 x + p_3$ & $[-0.2,0.2]$ & $[-2.0,2.0]$ & $[-3.0,3.0]$ & $-$\\
Cubic & $p_1 x^3 + p_2 x^2 +p_3 x + p_4$ & $[-0.1,0.1]$ & $[-0.2,0.2]$ & $[-2.0,2.0]$ & $[-3.0,3.0]$\\

\hline
\end{tabular}
\end{table*}

\subsubsection{Datasets for regression and classification}
For toy regression, to compare with \cite{hsml,maml}. We follow the task design from four function families. They are (1) Sinusoids (2)Line (3) Cubic (4) Quadratic. See Table~\ref{tab:regression_definition} for details of function definition. We also take MSE as the final evaluation metric. 

As for image few-shot classification, we use the widely accepted dataset,i.e. MiniImageNet, as benchmark to validate the effectiveness. MiniImageNt is a larger benchmark containing 100 classes randomly chosen from ImageNet  ILSVRC-2012 challenge \cite{imagenet} with 600 images of size 84×84 pixels per class. It is split into 64 base classes, 16 validation classes and 20 novel classes. 

\subsubsection{Models for regression and classification}
Following \cite{maml,hsml}, we take 2 hidden layer mlp and 4-conv network as backbones for toy regression and image classification respectively. The related hyperparameters are listed in Table~\ref{tab:hyperparameters}.

\begin{table*}
\centering
\caption{Hyperparameters of the experiments.}
\label{tab:hyperparameters}
\begin{tabular}[c]{c|c|c}
\hline
Hyperparameter  & Toy Regression  &  MiniImageNet\\
\hline
task batch size& 32  & 32\\
Inner loop learning rate & 0.001 &  0.03\\
Outer loop learning rate & 0.001 & 0.001 \\
Inner loop adaptation steps & 5 & 5 \\
epoch for pretraining & 15000 & 20000 \\
epoch for meta-train & 5000 & 10000 \\
K selected for k-means & 4 & 3 \\

\hline
\end{tabular}
\end{table*}

\begin{table*}
\centering
\caption{Performance of MSE $\pm$  95$\%$ confidence intervals on toy
regression tasks, averaged over 4,000 tasks. Both 5-shot and 10-
shot results are reported.}
\label{tab:toy_regression}
\begin{tabular}[c]{c|c|c}
\hline
Model  & 5-shot  &  10-shot\\
\hline
MAML& 2.205 $\pm$ 0.121  & 0.761 $\pm$ 0.068\\
Meta-SGD & 2.053 $\pm$ 0.117  & 0.836 $\pm$ 0.065 \\
MT-net & 2.435 $\pm$ 0.130 & 0.967 $\pm$ 0.056 \\
HSML & 0.856 $\pm$ 0.073 & 0.161 $\pm$ 0.021 \\
\hline
EEML & \bfseries 0.765 $\pm$ 0.140 & \bfseries 0.151 $\pm$ 0.032 \\

\hline
\end{tabular}
\end{table*}

\subsection{Toy Regression}
We compare our methods with recent state-of-the-art works in meta-learning, such as MAML, Meta-SGD, HSML. We can see from Table~\ref{tab:toy_regression} that our method achieves better results which indicates the effectiveness of our ensemble mechanism. 

\begin{table*}
\centering
\caption{Accuracy of image few-shot classification problem on MiniImageNet. Comparing with gradient-based meta-learning on 1-shot and 5-shot cases with 4-conv model as backbone. The results are with $\pm$  95$\%$ confidence intervals.}
\label{tab:image_classification}
\begin{tabular}[c]{c|c|c}
\hline
\multirow{2}{*}{Model}  & \multicolumn{2}{c}{MiniImageNet}  \\
\cline{2-3}
&1-shot  &  5-shot\\
\hline
MAML& 48.07 $\pm$ 1.75  & 63.15 $\pm$ 0.91\\
Meta-SGD &  50.47 $\pm$ 1.87 & 64.03 $\pm$ 0.94 \\
ANIL & 46.70 $\pm$ 0.40 & 61.50 $\pm$ 0.50 \\
BOIL & 49.61 $\pm$ 0.16 & 66.45 $\pm$ 0.37 \\
MT-net & 51.70 $\pm$ 1.84 & $-$ \\
sparse-MAML & 50.35 $\pm$ 0.39 & 67.03 $\pm$ 0.74 \\
HSML & 50.38 $\pm$ 1.85 & $-$ \\
\hline
EEML & \bfseries 52.42 $\pm$ 1.75   & \bfseries 68.40 $\pm$ 0.95   \\

\hline
\end{tabular}
\end{table*}

\subsection{Image Few-shot Classification}
We keep comparing with recent SOTA methods in much more complex images classification problem. As the Table~\ref{tab:image_classification} show, our method improves at least $\%4$ for original MAML, which easily outperform recent SOTA like Meta-SGD, MT-net, HSML. Both 1-shot and 5-shot cases earn stable improvement which validates the effectiveness of our method.

\section{Conclusion}
In this paper, we propose a novel ensemble embedded meta-learning(EEML) algorithm, and comparing with recent meta-learning on both toy regression and image few-shot classification tasks. The experiment results show that our method possesses great potential not only for simple task like function regression, but also complex task like image classification. Through replacing the single dictatorial expert with an cluster based ensemble voting mechanism, our method earn easily improvement comparing with recent state-of-the-arts. 

Although our method already earns explicit effects on two benchmarks, the potential of combining ensemble learning with meta-learning still calls for more exploration. And in reality employment case, the hyperparameters should be searched carefully to gain better performance.

%
%
%
\bibliographystyle{splncs04}
\bibliography{reference}

\end{document}